\documentclass[11pt,a4paper]{article}
\usepackage[hyperref]{acl2017}
\usepackage{times}
\usepackage{latexsym}

\usepackage{graphicx}
\usepackage{amsfonts}       
\usepackage{amsmath}       
\usepackage{amssymb}       
\usepackage{color}
\newtheorem{definition}{Definition}

\usepackage{url}

\aclfinalcopy 
\def\aclpaperid{61} 

\title{Does the Geometry of Word Embeddings Help Document Classification? \\
  A Case Study on Persistent Homology Based Representations}

\author{Paul Michel\thanks{*The indicated authors contributed equally to this work.} \\
  Carnegie Mellon University \\
  {\tt pmichel1@cs.cmu.edu} \\\And
  Abhilasha Ravichander\footnotemark[1] \\
  Carnegie Mellon University \\	
  {\tt aravicha@cs.cmu.edu} \\\And
  Shruti Rijhwani\footnotemark[1] \\
  Carnegie Mellon University \\
  {\tt srijhwan@cs.cmu.edu} \\}

\date{}

\begin{document}
\maketitle
\begin{abstract}
We investigate the pertinence of methods from algebraic topology for text data analysis. These methods enable the development of mathematically-principled isometric-invariant mappings from a set of vectors to a document embedding, which is stable with respect to the geometry of the document in the selected metric space. 
In this work, we evaluate the utility of these topology-based document representations in traditional NLP tasks, specifically document clustering and sentiment classification.
We find that the embeddings do not benefit text analysis. In fact, performance is worse than simple techniques like \textit{tf-idf}, indicating that the geometry of the document does not provide enough variability for classification on the basis of topic or sentiment in the chosen datasets.
\end{abstract}

\section{Introduction}

Given a embedding model mapping words to $n$ dimensional vectors, every document can be represented as a finite subset of $\mathbb{R}^n$. Comparing documents then amounts to comparing such subsets. While previous work shows that the Earth Mover's Distance \citep{kusner2015word} or distance between the weighted average of word vectors \citep{arora2017simple} provides information that is useful for classification tasks, we wish to go a step further and investigate whether useful information can also be found in the `shape' of a document in word embedding space.

Persistent homology is a tool from algebraic topology used to compute topological signatures (called  \emph{persistence diagrams}) on compact metric spaces. These have the property of being stable with respect to the Gromov-Haussdorff distance~\cite{gromov1981structures}. In other words, compact metric spaces that are close, up to an isometry, will have similar embeddings. In this work, we examine the utility of such embeddings in text classification tasks. To the best of our knowledge, no previous work has been performed on using topological representations for traditional NLP tasks, nor has any comparison been made with state-of-the-art approaches.


We begin by considering a document as the set of its word vectors, generated with a pretrained word embedding model. These form the metric space on which we build persistence diagrams, using Euclidean distance as the distance measure. The diagrams are a representation of the document's geometry in the metric space. We then perform clustering on the Twenty Newsgroups dataset with the features extracted from the persistence diagram. We also evaluate the method on sentiment classification tasks, using the Cornell Sentence Polarity (CSP) \cite{PangLee:05a} and IMDb movie review datasets \cite{maas-EtAl}.

As suggested by \citet{zhu13}, we posit that the information about the intrinsic geometry of documents, found in the persistence diagrams, might yield information that our classifier can leverage, either on its own or in combination with other representations. The primary objective of our work is to empirically evaluate these representations in the case of sentiment and topic classification, and assess their usefulness for real-world tasks.
\section{Method}
\label{sec:method}

\subsection{Word embeddings}

As a first step we compute word vectors for each document in our corpus using a word2vec \cite{mikolov} model trained on the Google News dataset\footnote{https://code.google.com/archive/p/word2vec/}.
In addition to being a widely used word embedding technique, word2vec has been known to exhibit interesting linear properties with respect to analogies \cite{mikolov}, which hints at rich semantic structure.

\subsection{Gromov-Haussdorff Distance}

Given a dictionary of word vectors of dimension $n$, we can represent any document as a finite subset of $\mathbb R ^ n$. The \textit{Haussdorff distance} gives us a way to evaluate the distance between two such sets. More precisely, the Haussdorff distance $d_H$ between two finite subsets $A,B$ of $\mathbb{R}^n$ is defined as:
\[d_{H}(A,B)=\max(\sup_{a\in A}d(a,B),\sup_{b\in B}d(b,A))\]
where $d(x,Y)=\inf_{y\in Y}\Vert x - y\Vert_2$ is the distance of point $x$ from set $Y$.

However, this distance is sensitive to translations and other isometric\footnote{$f:\mathbb{R}^n\longrightarrow\mathbb{R}^n$ is \emph{isometric} if it is distance preserving, ie $\forall x,y\in\mathbb{R}^n,\Vert f(x)-f(y)\Vert_2=\Vert x - y\Vert_2$. Rotations, translations and reflections are examples of (linear) isometries.} transformations. Hence, a more natural metric is the \textbf{Gromov-Haussdorff distance} \cite{gromov1981structures}, simply defined as
\[d_{GH}(A,B)=\inf_{f\in E_n}d_H(A,f(B))\]
where $E_n$ is the set of all isometries of $\mathbb{R}^n$.

\begin{figure}[tb]
\begin{center}\includegraphics[width=0.9\columnwidth]{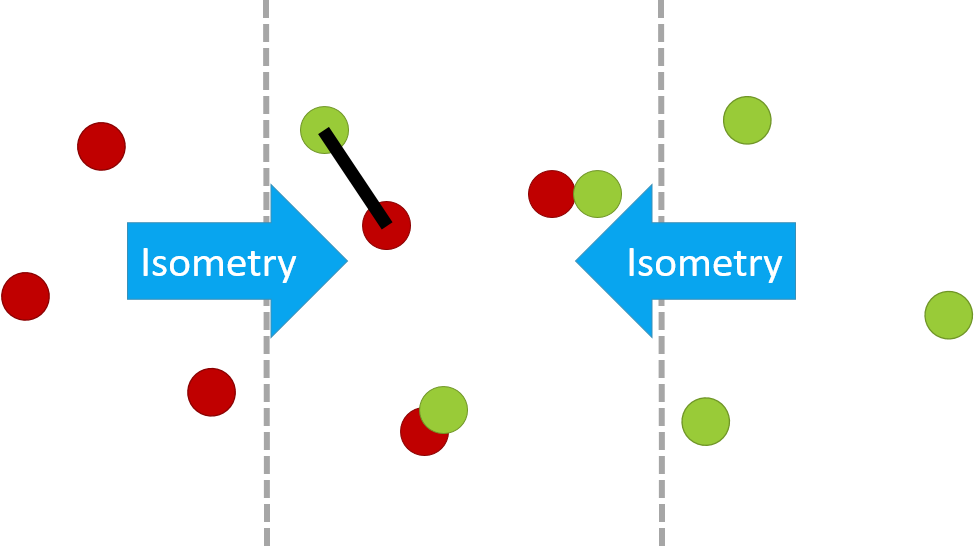}
\caption{Gromov-Haussdorff distance between two sets (red, green). The black bar represents the actual distance (given that the isometric embedding is optimal).}
\label{fig:ghdist}
\end{center}
\end{figure}

Figure \ref{fig:ghdist} provides an example of practical Gromov-Haussdorff (GH) distance computation between two sets of three points each. Both sets are embedded in $\mathbb{R}^2$ (middle panel) using isometries i.e the distance between points in each set is conserved. The Haussdorff distance between the two embedded sets corresponds to the length of the black segment. The GH distance is the minimum Haussdorff distance under all possible isometric embeddings.

We want to compare documents based on their intrinsic geometric properties. Intuitively, the GH distance measures how far two sets are from being isometric. This allows us to define the geometry of a document more precisely:

\begin{definition}[Document Geometry]
We say that two documents A, B have the same \emph{geometry} if $d_{GH}(A,B)=0$, ie if they are the same up to an isometry.
\end{definition}

Mathematically speaking, this amounts to defining the geometry of a document as its equivalence class under the equivalence relation induced by the GH distance on the set of all documents. 

\subparagraph{Comparison to the Earth Mover Distance}: \citet{kusner2015word} proposed a new method for computing a distance between documents based on an instance of the Earth Mover Distance \cite{rubner1998metric} called Word Mover Distance (WMD). While WMD quantifies the total cost of matching all words of one document to another, the GH distance is the cost, up to an isometry, of the worst-case matching.

\subsection{Persistence diagrams}
\label{sec:pers}

Efficiently computing the GH distance is still an open problem despite a lot of recent work in this area \cite{memoli2005theoretical,bronstein2006efficient,memoli2007use,agarwal2015computing}.

Fortunately, \citet{carriere2015stable} provides us with a way to derive a signature which is stable with respect to the GH distance. More specifically, given a finite point cloud $A\subset \mathbb{R}^n$, the persistence diagram of the Vietori-Rips filtration on $A$, $Dg(A)$, can be computed. This approach is inspired by persistent homology, a subfield of algebraic topology.

The rigorous definition of these notions is not the crux of this paper and we will only present them informally. The curious reader is invited to refer to \citet{zhu13} for a short introduction. More details are in \citet{delfinado1995incremental,edelsbrunner2002topological,robins1999towards}.

A persistence diagram is a scatter plot of 2-D points representing the appearance and disappearance of geometric features\footnote{such as connected components, holes or empty hulls} under varying resolutions. This can be imagined as replacing each point by a sphere of increasing radius.

We use the procedure described in \citet{carriere2015stable} to derive fixed-sized vectors from persistence diagrams. These vectors have the following property: if $A$ and $B$ are two finite subsets of $\mathbb{R}^n$, $Dg(A)$ and $Dg(B)$ are their persistence diagrams, $N=\max(\vert Dg(A)\vert,\vert Dg(B)\vert$) and $V_A,V_B\in \mathbb{R}^\frac{N(N-1)}{2}$, then 
\[\Vert V_A-V_B\Vert_2\leqslant \sqrt{2N(N-1)} d_{GH}(A,B)\]
In other words, the resulting signatures $V_A$ and $V_B$ are stable with respect to the GH distance. The size of the vectors are dependent on the underlying sets $A$ and $B$. However, as is argued in \citet{carriere2015stable}, we can truncate the vectors to a dimension fixed across our dataset while preserving the stability property (albeit losing some of the representative ability of the signatures).

\section{Experiments}
\label{sec:exp}

\subsection{Experiments}

\begin{figure}[tb]
\begin{center}\includegraphics[width=\columnwidth]{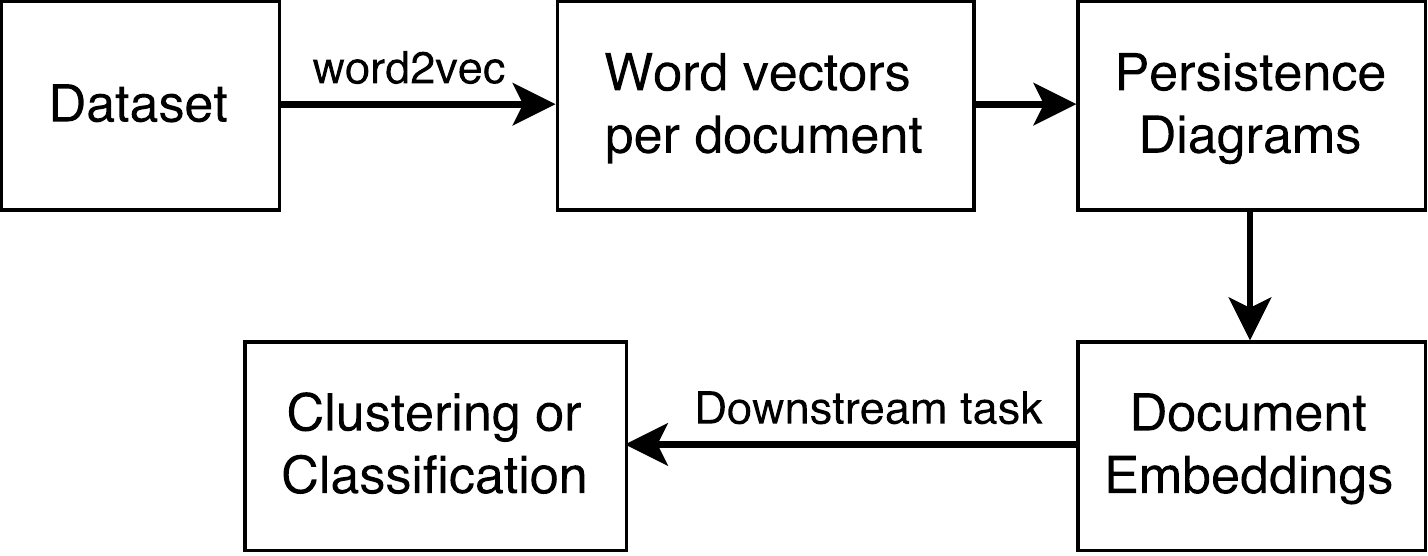}
\caption{Method Pipeline}
\label{simcom}
\end{center}
\end{figure}

The pipeline for our experiments is shown in Figure~\ref{simcom}. In order to build a persistence diagram, we convert each document to the set of its word vectors. We then use Dionysus \cite{dyonisus}, a C++ library for computing persistence diagrams, and form the signatures described in \ref{sec:pers}. We will subsequently refer to these diagrams as Persistent Homology (\textsc{Ph}) embeddings. Once we have the embeddings for each document, they can be used as input to standard clustering or classification algorithms. 

As a baseline document representation, we use the average of the word vectors for that document (subsequently called \textsc{Aw2v} embeddings).

For clustering, we experiment with K-means and Gaussian Mixture Models (GMM) on a subset\footnote{alt.atheism, sci.space and talk.religion.misc categories} of the Twenty Newsgroups dataset. The subset was selected to ensure that most documents are from related topics, making clustering non-trivial, and the documents are of reasonable length to compute the representation.

For classification, we perform both sentence-level and document-level binary sentiment classification using logistic regression on the CSP and IMDb corpora respectively. 

\begin{figure*}[tb]
\centering\includegraphics[width=0.8\textwidth]{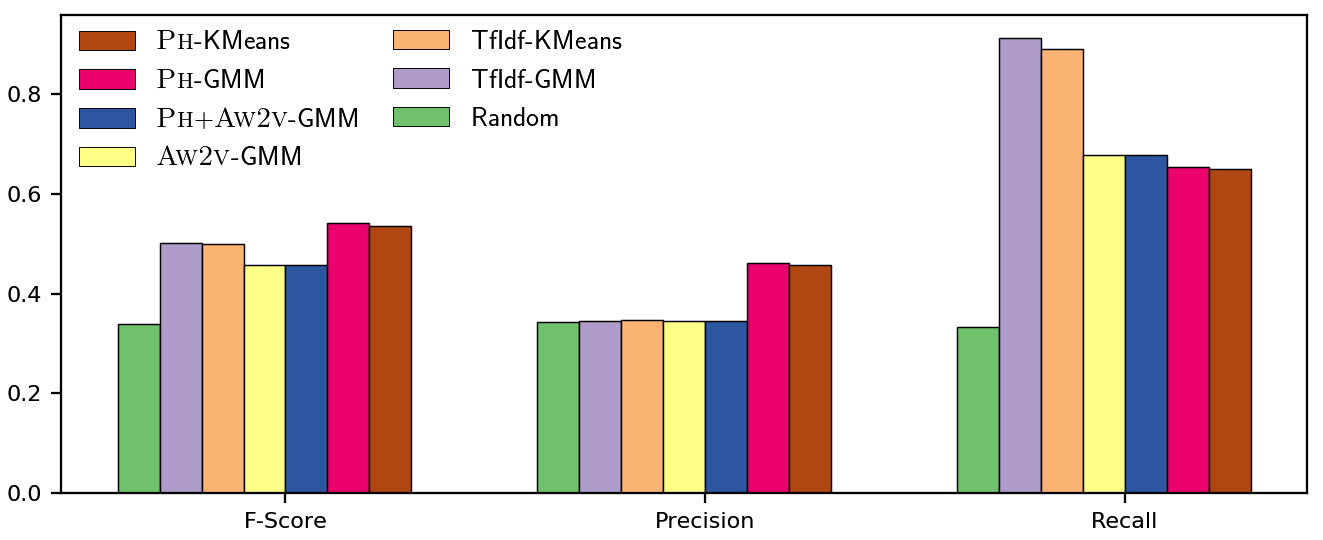}
\caption{Results for clustering on 3 subclasses of the Twenty Newsgroups dataset}
\label{results}
\end{figure*}
\section{Results}
\label{sec:results}

\subsection{Hyper-parameters}

Our method depends on very few hyper-parameters. Our main choices are listed below.

\paragraph{Choice of distance} We experimented with both euclidean distance and cosine similarity (angular distance). After preliminary experiments, we determined that both performed equally and hence, we only report results with the euclidean distance.
\paragraph{Persistence diagram computation} The hyper-parameters of the diagram computation are monotonic and mostly control the degree of approximation. We set them to the highest values that allowed our experiment to run in reasonable time\footnote{Selected such that the computation of the diagram of the longest file in the training data took less than 10 minutes.}.

\subsection{Document Clustering}

We perform clustering experiments with the baseline document features (\textsc{Aw2v}), \textit{tf-idf} and our \textsc{Ph} signatures. Figure \ref{results} shows the B-Cubed precision, recall and F1-Score of each method (metrics as defined in \citet{amigo2009comparison}). To further assess the utility of \textsc{Ph} embeddings, we concatenate them with \textsc{Aw2v} to obtain a third representation, \textsc{Aw2v+Ph}.

With GMM and \textsc{Aw2v+Ph}, the F1-Score of clustering is $0.499$. In terms of F1 and precision, we see that \textit{tf-idf} representations perform better than \textsc{Ph}, for reasons that we will discuss in later sections. In terms of recall, \textsc{Ph} as well as \textsc{Aw2v} perform fairly well. Importantly, we see that all the metrics for \textsc{Ph} are significantly above the random baseline, indicating that some valuable information is contained in them. 

\subsection{Sentiment Classification}
\subsubsection{Sentence-Level Sentiment Analysis}
We evaluate our method on the CSP dataset\footnote{For lack of a canonical split, we use a random 10\% of the dataset as a test set}. The results are presented in Table \ref{tab:csp}. For comparison, we provide results for one of the state of the art models, a CNN-based sentence classifier \cite{Kim14convolutionalneural}. We observe that by themselves, \textsc{Ph} embeddings are not useful at predicting the sentiment of each sentence. \textsc{Aw2v} gives reasonable performance in this task, but combining the two representations does not impact the accuracy at all.

\subsubsection{Document-Level Sentiment Analysis}
We perform document-level binary sentiment classification on the IMDb Movie Reviews Dataset \cite{maas-EtAl}. We use sentence vectors in this experiment, each of which is the average of the word vectors in that sentence. The results are presented in Table \ref{tab:imdb}. We compare our results with the paragraph-vector approach \cite{MikolovDoc2Vec}. We observe that \textsc{Ph} embeddings perform poorly on this dataset. Similar to the CSP dataset, \textsc{Aw2v} embeddings give acceptable results. The combined representation performs slightly better, but not by a margin of significance.

\begin{table}[tb]
\centering
\begin{tabular}{|l|c|}
\hline
Model                              & Accuracy \\ \hline
CNN Non-Static    & 81.5\%   \\ 
\textsc{Ph} + LogReg                      & 53.19\%  \\ 
\textsc{Aw2v} + LogReg             & 77.13\%  \\ 
\textsc{Aw2v} + \textsc{Ph} + LogReg  & 77.13\%  \\ \hline
\end{tabular}
\caption{\label{tab:csp}Performance on the CSP dataset}
\end{table}

\begin{table}[tb]
\centering
\begin{tabular}{|l|c|}
\hline
Model                                   & Accuracy \\ \hline
Paragraph Vector  & 92.58\%  \\ 
\textsc{Ph} + LogReg          & 53.16\%  \\ 
\textsc{Aw2v} + LogReg              & 82.94\%  \\ 
\textsc{Aw2v} + \textsc{Ph} + LogReg   & 83.08\%  \\ \hline
\end{tabular}
\caption{\label{tab:imdb}Performance on the IMDb dataset}

\end{table} 

\section{Discussion and Analysis}
As seen in Figure~\ref{results}, the \textsc{Ph} representation does not outperform \textit{tf-idf} or \textsc{Aw2v}, and in fact often doesn't perform much better than chance. 

One possible reason is linked to the nature of our datasets: the computation of the persistence diagram is very sensitive to the size of the documents. The geometry of small documents, where the number of words is negligible with respect to the dimensionality of the word vectors, is not very rich. The resulting topological signatures are very sparse, which is a problem for CSP as well as documents in IMDb and Twenty Newsgroups that contain only one line. 
On the opposite side of the spectrum, persistence diagrams are intractable to compute without down-sampling for very long documents (which in turn negatively impacts the representation of smaller documents).

We performed an additional experiment on a subset of the IMDb corpus that only contained documents of reasonable length, but obtained similar results. This indicates that the poor performance of \textsc{Ph} representations, even when combined with other features (\textsc{Aw2v}), cannot be explained only by limitations of the data.

These observations lead to the conclusion that, for these datasets, the intrinsic geometry of documents in the word2vec semantic space does not help text classification tasks.

\section{Related Work}
\label{sec:related}

Learning distributed representations of sentences or documents for downstream classification and information retrieval tasks has received recent attention owing to their utility in several applications, be it representations trained on the sentence/paragraph level \citet{MikolovDoc2Vec,kiros2015skip} or  purely word vector based methods \citet{arora2017simple}.

Document classification and clustering \cite{willett1988recent, hotho2005brief,steinbach2000comparison, huang2008similarity, Xu, kuang2015nonnegative, miller2016unsupervised} and sentiment classification \cite{Nakagawa, Kim14convolutionalneural,WangLiang} are relatively well studied.

Topological data analysis has been used for various tasks such as 3D shapes classification \cite{chazal2009gromov} or protein structure analysis \cite{xia2014persistent}. However, such techniques have not been used in NLP, primarily because the theory is inaccessible and suitable applications are scarce. \citet{zhu13} offers an introduction to using persistent homology in NLP, by creating representations of nursery-rhymes and novels, as well as highlights structural differences between child and adolescent writing. However, these techniques have not been applied to core NLP tasks. 

\section{Conclusion}

Based on our experiments, using persistence diagrams for text representation does not seem to positively contribute to document clustering and sentiment classification tasks. There are certainly merits to the method, specifically its strong mathematical foundation and its domain-independent, unsupervised nature. Theoretically, algebraic topology has the ability to capture structural context, and this could potentially benefit syntax-based NLP tasks such as parsing. We plan to investigate this connection in the future.  

\section*{Acknowledgments}
This work was supported in part by the Defense Advanced Research Projects Agency (DARPA) Information Innovation Office (I2O) under the Low Resource Languages for Emergent Incidents
(LORELEI) program issued by DARPA/I2O under Contract No. HR0011-15-C-0114. The views expressed are those of the authors and do not reflect the official policy or position of the Department of Defense or the U.S. Government.

We are grateful to Matt Gormley, Hyun Ah Song, Shivani Poddar and Hai Pham for their suggestions on the writing of this paper as well as to Steve Oudot for pointing us to helpful references. We would also like to thank the anonymous ACL reviewers for their valuable suggestions.





\bibliography{acl2017}

\begin{thebibliography}{}
\expandafter\ifx\csname natexlab\endcsname\relax\def\natexlab#1{#1}\fi

\bibitem[{Agarwal et~al.(2015)Agarwal, Fox, Nath, Sidiropoulos, and
  Wang}]{agarwal2015computing}
Pankaj~K Agarwal, Kyle Fox, Abhinandan Nath, Anastasios Sidiropoulos, and Yusu
  Wang. 2015.
\newblock Computing the gromov-hausdorff distance for metric trees.
\newblock In {\em International Symposium on Algorithms and Computation\/}.
  Springer, pages 529--540.

\bibitem[{Amig{\'o} et~al.(2009)Amig{\'o}, Gonzalo, Artiles, and
  Verdejo}]{amigo2009comparison}
Enrique Amig{\'o}, Julio Gonzalo, Javier Artiles, and Felisa Verdejo. 2009.
\newblock A comparison of extrinsic clustering evaluation metrics based on
  formal constraints.
\newblock {\em Information retrieval\/} 12(4):461--486.

\bibitem[{Arora et~al.(2017)Arora, Liang, and Ma}]{arora2017simple}
Sanjeev Arora, Yingyu Liang, and Tengyu Ma. 2017.
\newblock A simple but tough-to-beat baseline for sentence embeddings.
\newblock In {\em International Conference on Learning Representations. To
  Appear\/}.

\bibitem[{Bronstein et~al.(2006)Bronstein, Bronstein, and
  Kimmel}]{bronstein2006efficient}
Alexander~M Bronstein, Michael~M Bronstein, and Ron Kimmel. 2006.
\newblock Efficient computation of isometry-invariant distances between
  surfaces.
\newblock {\em SIAM Journal on Scientific Computing\/} 28(5):1812--1836.

\bibitem[{Carri{\`e}re et~al.(2015)Carri{\`e}re, Oudot, and
  Ovsjanikov}]{carriere2015stable}
Mathieu Carri{\`e}re, Steve~Y Oudot, and Maks Ovsjanikov. 2015.
\newblock Stable topological signatures for points on 3d shapes.
\newblock In {\em Computer Graphics Forum\/}. Wiley Online Library, volume~34,
  pages 1--12.

\bibitem[{Chazal et~al.(2009)Chazal, Cohen-Steiner, Guibas, M{\'e}moli, and
  Oudot}]{chazal2009gromov}
Fr{\'e}d{\'e}ric Chazal, David Cohen-Steiner, Leonidas~J Guibas, Facundo
  M{\'e}moli, and Steve~Y Oudot. 2009.
\newblock Gromov-hausdorff stable signatures for shapes using persistence.
\newblock In {\em Computer Graphics Forum\/}. Wiley Online Library, volume~28,
  pages 1393--1403.

\bibitem[{Delfinado and Edelsbrunner(1995)}]{delfinado1995incremental}
Cecil Jose~A Delfinado and Herbert Edelsbrunner. 1995.
\newblock An incremental algorithm for betti numbers of simplicial complexes on
  the 3-sphere.
\newblock {\em Computer Aided Geometric Design\/} 12(7):771--784.

\bibitem[{Edelsbrunner et~al.(2002)Edelsbrunner, Letscher, and
  Zomorodian}]{edelsbrunner2002topological}
Herbert Edelsbrunner, David Letscher, and Afra Zomorodian. 2002.
\newblock Topological persistence and simplification.
\newblock {\em Discrete and Computational Geometry\/} 28(4):511--533.

\bibitem[{Gromov et~al.(1981)Gromov, Lafontaine, and
  Pansu}]{gromov1981structures}
Mikhael Gromov, Jacques Lafontaine, and Pierre Pansu. 1981.
\newblock Structures m{\'e}triques pour les vari{\'e}t{\'e}s riemanniennes .

\bibitem[{Hotho et~al.(2005)Hotho, N{\"u}rnberger, and
  Paa{\ss}}]{hotho2005brief}
Andreas Hotho, Andreas N{\"u}rnberger, and Gerhard Paa{\ss}. 2005.
\newblock A brief survey of text mining.
\newblock In {\em Ldv Forum\/}.

\bibitem[{Huang(2008)}]{huang2008similarity}
Anna Huang. 2008.
\newblock Similarity measures for text document clustering.
\newblock In {\em Proceedings of the sixth new zealand computer science
  research student conference (NZCSRSC2008), Christchurch, New Zealand\/}.
  pages 49--56.

\bibitem[{Kim(2014)}]{Kim14convolutionalneural}
Yoon Kim. 2014.
\newblock Convolutional neural networks for sentence classification.
\newblock In {\em In EMNLP\/}.

\bibitem[{Kiros et~al.(2015)Kiros, Zhu, Salakhutdinov, Zemel, Urtasun,
  Torralba, and Fidler}]{kiros2015skip}
Ryan Kiros, Yukun Zhu, Ruslan~R Salakhutdinov, Richard Zemel, Raquel Urtasun,
  Antonio Torralba, and Sanja Fidler. 2015.
\newblock Skip-thought vectors.
\newblock In {\em Advances in neural information processing systems\/}. pages
  3294--3302.

\bibitem[{Kuang et~al.(2015)Kuang, Choo, and Park}]{kuang2015nonnegative}
Da~Kuang, Jaegul Choo, and Haesun Park. 2015.
\newblock Nonnegative matrix factorization for interactive topic modeling and
  document clustering.
\newblock In {\em Partitional Clustering Algorithms\/}, Springer, pages
  215--243.

\bibitem[{Kusner et~al.(2015)Kusner, Sun, Kolkin, Weinberger
  et~al.}]{kusner2015word}
Matt~J Kusner, Yu~Sun, Nicholas~I Kolkin, Kilian~Q Weinberger, et~al. 2015.
\newblock From word embeddings to document distances.
\newblock In {\em ICML\/}. volume~15, pages 957--966.

\bibitem[{Le and Mikolov(2014)}]{MikolovDoc2Vec}
Quoc~V. Le and Tomas Mikolov. 2014.
\newblock Distributed representations of sentences and documents.
\newblock In {\em Proceedings of the 31th International Conference on Machine
  Learning, {ICML} 2014, Beijing, China, 21-26 June 2014\/}. pages 1188--1196.

\bibitem[{Maas et~al.(2011)Maas, Daly, Pham, Huang, Ng, and Potts}]{maas-EtAl}
Andrew~L. Maas, Raymond~E. Daly, Peter~T. Pham, Dan Huang, Andrew~Y. Ng, and
  Christopher Potts. 2011.
\newblock \href{http://www.aclweb.org/anthology/P11-1015}{Learning word vectors
  for sentiment analysis}.
\newblock In {\em Proceedings of the 49th Annual Meeting of the Association for
  Computational Linguistics: Human Language Technologies\/}. Association for
  Computational Linguistics, Portland, Oregon, USA, pages 142--150.
\newblock
  \href{http://www.aclweb.org/anthology/P11-1015}{http://www.aclweb.org/anthology/P11-1015}.

\bibitem[{M{\'e}moli(2007)}]{memoli2007use}
Facundo M{\'e}moli. 2007.
\newblock On the use of gromov-hausdorff distances for shape comparison .

\bibitem[{M{\'e}moli and Sapiro(2005)}]{memoli2005theoretical}
Facundo M{\'e}moli and Guillermo Sapiro. 2005.
\newblock A theoretical and computational framework for isometry invariant
  recognition of point cloud data.
\newblock {\em Foundations of Computational Mathematics\/} 5(3):313--347.

\bibitem[{Mikolov et~al.(2013)Mikolov, Sutskever, Chen, Corrado, and
  Dean}]{mikolov}
Tomas Mikolov, Ilya Sutskever, Kai Chen, Greg~S Corrado, and Jeff Dean. 2013.
\newblock Distributed representations of words and phrases and their
  compositionality.
\newblock In {\em Advances in neural information processing systems\/}. pages
  3111--3119.

\bibitem[{Miller et~al.(2016)Miller, Dligach, and
  Savova}]{miller2016unsupervised}
Timothy~A Miller, Dmitriy Dligach, and Guergana~K Savova. 2016.
\newblock Unsupervised document classification with informed topic models.
\newblock {\em ACL\/} .

\bibitem[{Morozov(2008--2016)}]{dyonisus}
Dmitriy Morozov. 2008--2016.
\newblock {Dyonisus} : a c++ library for computing persistent homology.
\newblock \url{http://mrzv.org/software/dionysus/}.

\bibitem[{Nakagawa et~al.(2010)Nakagawa, Inui, and Kurohashi}]{Nakagawa}
Tetsuji Nakagawa, Kentaro Inui, and Sadao Kurohashi. 2010.
\newblock \href{http://dl.acm.org/citation.cfm?id=1857999.1858119}{Dependency
  tree-based sentiment classification using crfs with hidden variables}.
\newblock In {\em Human Language Technologies: The 2010 Annual Conference of
  the North American Chapter of the Association for Computational
  Linguistics\/}. Association for Computational Linguistics, Stroudsburg, PA,
  USA, HLT '10, pages 786--794.
\newblock
  \href{http://dl.acm.org/citation.cfm?id=1857999.1858119}{http://dl.acm.org/citation.cfm?id=1857999.1858119}.

\bibitem[{Pang and Lee(2005)}]{PangLee:05a}
Bo~Pang and Lillian Lee. 2005.
\newblock Seeing stars: Exploiting class relationships for sentiment
  categorization with respect to rating scales.
\newblock In {\em Proceedings of the ACL\/}.

\bibitem[{Robins(1999)}]{robins1999towards}
Vanessa Robins. 1999.
\newblock Towards computing homology from finite approximations.
\newblock In {\em Topology proceedings\/}. volume~24, pages 503--532.

\bibitem[{Rubner et~al.(1998)Rubner, Tomasi, and Guibas}]{rubner1998metric}
Yossi Rubner, Carlo Tomasi, and Leonidas~J Guibas. 1998.
\newblock A metric for distributions with applications to image databases.
\newblock In {\em Computer Vision, 1998. Sixth International Conference on\/}.
  IEEE, pages 59--66.

\bibitem[{Steinbach et~al.(2000)Steinbach, Karypis, Kumar
  et~al.}]{steinbach2000comparison}
Michael Steinbach, George Karypis, Vipin Kumar, et~al. 2000.
\newblock A comparison of document clustering techniques.
\newblock In {\em KDD workshop on text mining\/}.

\bibitem[{Wang and Manning(2012)}]{WangLiang}
Sida Wang and Christopher~D. Manning. 2012.
\newblock \href{http://dl.acm.org/citation.cfm?id=2390665.2390688}{Baselines
  and bigrams: Simple, good sentiment and topic classification}.
\newblock In {\em Proceedings of the 50th Annual Meeting of the Association for
  Computational Linguistics: Short Papers - Volume 2\/}. Association for
  Computational Linguistics, Stroudsburg, PA, USA, ACL '12, pages 90--94.
\newblock
  \href{http://dl.acm.org/citation.cfm?id=2390665.2390688}{http://dl.acm.org/citation.cfm?id=2390665.2390688}.

\bibitem[{Willett(1988)}]{willett1988recent}
Peter Willett. 1988.
\newblock Recent trends in hierarchic document clustering: a critical review.
\newblock {\em Information Processing \& Management\/} 24(5):577--597.

\bibitem[{Xia and Wei(2014)}]{xia2014persistent}
Kelin Xia and Guo-Wei Wei. 2014.
\newblock Persistent homology analysis of protein structure, flexibility, and
  folding.
\newblock {\em International journal for numerical methods in biomedical
  engineering\/} 30(8):814--844.

\bibitem[{Xu and Gong(2004)}]{Xu}
Wei Xu and Yihong Gong. 2004.
\newblock \href{https://doi.org/10.1145/1008992.1009029}{Document clustering by
  concept factorization}.
\newblock In {\em Proceedings of the 27th Annual International ACM SIGIR
  Conference on Research and Development in Information Retrieval\/}. ACM, New
  York, NY, USA, SIGIR '04, pages 202--209.
\newblock
  \href{https://doi.org/10.1145/1008992.1009029}{https://doi.org/10.1145/1008992.1009029}.

\bibitem[{Zhu(2013)}]{zhu13}
Xiaojin Zhu. 2013.
\newblock Persistent homology: An introduction and a new text representation
  for natural language processing.
\newblock In {\em Proceedings of the 23rd International Joint Conference on
  Artificial Intelligence\/}.

\end{thebibliography}
\bibliographystyle{acl_natbib}

\appendix

\end{document}